\documentclass[letterpaper, 10 pt, conference]{ieeeconf}  
\IEEEoverridecommandlockouts                              

\usepackage{caption}
\usepackage{cite}
\usepackage{mathrsfs}
\usepackage{multirow}
\usepackage{amsmath}
\usepackage{amssymb}
\usepackage{setspace}
\usepackage{array}
\usepackage{subfigure}
\usepackage{mathtools}
\usepackage{algorithm}
\usepackage{algpseudocode}
\usepackage{float}
\usepackage{color}
\usepackage{xcolor}
\usepackage{hyperref}
\usepackage{siunitx}
\usepackage{stackengine}
\let\labelindent\relax
\usepackage{enumitem}
\usepackage{bm}

\usepackage[keeplastbox]{flushend}

\DeclareSIUnit\pixel{pixels}

\long\def\invis#1{}

\DeclareSIUnit\dBm{dBm}

\usepackage{makeidx}         
\usepackage{graphicx}        
\usepackage{multicol}        
\usepackage[bottom]{footmisc}

\title{\LARGE \bf
DeepSemanticHPPC: Hypothesis-based Planning over Uncertain Semantic Point Clouds
}

\author{Yutao Han$^{*}$, Hubert Lin$^{*}$, Jacopo Banfi$^{*}$, Kavita Bala, and Mark Campbell
\thanks{$\hspace{-6mm}^{*}$Equal contribution; order determined randomly.}
\thanks{\hspace{-6mm} All the authors are with Cornell University, Ithaca NY, USA {\tt\small \{yh675,hl2247,jb2639,mc288,kb97\}@cornell.edu}.}
\thanks{\hspace{-6mm} This work is funded by the ONR under the PERISCOPE MURI Grant N00014-17-1-2699.}
\thanks{\hspace{-6mm} Project page at \texttt{https://deepsemantichppc.github.io}}
}

\begin{document}

\clearpage
\thispagestyle{empty}
\onecolumn
\noindent \textcopyright 2020 IEEE. Personal use of this material is permitted.  Permission from IEEE must be obtained for all other uses, in any current or future media, including reprinting/republishing this material for advertising or promotional purposes, creating new collective works, for resale or redistribution to servers or lists, or reuse of any copyrighted component of this work in other works.

\bigskip

\noindent This is the author manuscript without publisher editing. An edited version of this manuscript will be published in proceedings of the 2020 International Conference on Robotics and Automation (ICRA). 






















\twocolumn
\clearpage

\maketitle
\thispagestyle{empty}
\pagestyle{empty}

\maketitle
\thispagestyle{empty}
\pagestyle{empty}

\begin{abstract}
Planning in unstructured environments is challenging -- it relies on sensing,
perception, scene reconstruction, and reasoning about various uncertainties. We
propose DeepSemanticHPPC, a novel uncertainty-aware hypothesis-based planner for
unstructured environments. Our algorithmic pipeline consists of: a deep Bayesian
neural network which segments surfaces with uncertainty estimates; a flexible
point cloud scene representation; a next-best-view planner which minimizes the
uncertainty of scene semantics using \emph{sparse} visual measurements; and a
hypothesis-based path planner that proposes multiple kinematically feasible
paths with evolving safety confidences given next-best-view measurements.
Our pipeline iteratively decreases semantic uncertainty
along planned paths, filtering out unsafe paths with high
confidence. We show that our framework plans safe paths in real-world
environments where existing path planners typically fail.

\end{abstract}

\section{Introduction}
\label{sec:intro}

Path planning for complex outdoor environments is challenging due to the
unstructured nature of environments that do not fall neatly into discretized
space. Moreover, different terrain surface types can be difficult to detect with
traditional sensing modalities. In indoor environments, a grid space
representation with lidar sensors is sufficient
\cite{LearningSubgoals,IvanovPlan,Lavalle}. Outdoor
environments exhibit complex geometries and surface
types, which are difficult--if not impossible--to differentiate using
just lidar data. Therefore, a more flexible scene representation, surface classification using computer vision techniques, and reasoning about scene uncertainties are necessary.

Previous work for outdoor planning has focused on classifying terrain and surface roughness using SVM classifiers
\cite{TerrainSVM,AgriTerrain}, neural networks \cite{LearningGT,OntheSpotTerrain},
and various other computer vision techniques \cite{ColorTerrain,VisTerrainLegged}. While these techniques can differentiate between simple terrain types, they do not model the inherent uncertainties and ambiguities in complex scenes which make it
difficult to differentiate between terrain types (e.g., an offroad robot
driving through a patch of grass with small rocks). Many current outdoor
planning approaches still rely on grid maps which do not model the complex
geometry of an outdoor scene (e.g., a field with irregular bumps and
rocks) \cite{PlanningImageSpace,FieldD}.
Recent work models outdoor maps for planning with a point cloud
\cite{krusi2017driving}, which is more flexible and suitable for
unstructured scenes; however, \cite{krusi2017driving} uses traditional lidar
sensing which cannot differentiate between different surface types as broadly
as computer vision.

\begin{figure}[t!]
\centering {\includegraphics[width=0.9\columnwidth]{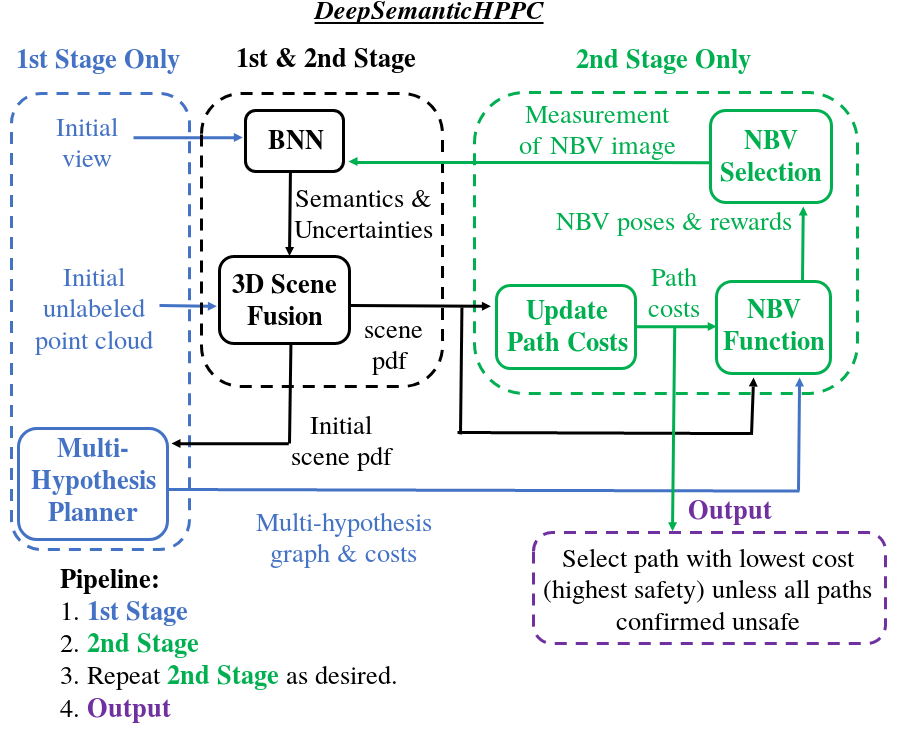}}
\caption{The DeepSemanticHPPC pipeline. (1) In the first stage, initial inputs
  generate a multi-hypothesis graph of possible paths. (2) In the
  second stage, the uncertainty in the scene is reduced and path costs are
  updated. (3) The second stage is repeated for a set number of iterations.
  This is terminated early if a safe path is confirmed or all paths are
  confirmed as unsafe. (4) A path is selected.  \label{fig:pipeline}}
  \vspace{-0.35in}
\end{figure}





In this paper we present DeepSemanticHPPC (Deep Semantic Hypothesis-based
Planner over Point Clouds), a novel algorithmic pipeline for planning over uncertain semantic point clouds, which leverages a Bayesian neural network (BNN) \cite{Gal2016Dropout, KendallUncert}
to extract principled estimates of segmentation uncertainty. This allows our framework 
to reason about ambiguous terrain as well as robustly handle false positive
detections by taking additional measurements to reduce semantic uncertainty in
the scene. However, each measurement is costly due to the computationally
expensive nature of Bayesian neural networks operating on a robotic platform
with limited computing power. Our planner hence employs next-best-view (NBV) techniques \cite{SEE,
LearntoScore, DaudNBV} to optimize for new measurements. DeepSemanticHPPC includes:

\begin{itemize}[leftmargin=*]

   \item the employment of a deep Bayesian neural network \cite{Gal2016Dropout, KendallUncert} to obtain surface and obstacle semantics with uncertainty estimates for
    unstructured outdoor environments;

   \item a flexible point cloud scene representation;

   \item a next-best-view planner which minimizes the uncertainty of
     terrain semantics using \emph{sparse} visual measurements;

  \item a hypothesis-based path planner (extending \cite{krusi2017driving})
    that proposes multiple kinematically feasible paths with evolving safety
    confidences given the NBV measurements.
  
\end{itemize}

Experimental results with real environments show that our pipeline 
plans safe paths in real-world environments where existing path planners
typically fail. Fig. \ref{fig:pipeline} illustrates DeepSemanticHPPC. In the first stage, a multi-hypothesis planner generates multiple hypotheses of possible safe paths given a scene belief.
In the second stage, a NBV function calculates NBV poses and associated rewards. These poses and rewards are input to a
NBV selection block which selects the best \textit{feasible} NBV. A BNN extracts semantic segmentations and associated uncertainties
from the NBV measurement,
which are used to generate a new scene belief. The new scene belief reduces
hypothesis uncertainty, and the second stage is repeated for a set number of iterations.
Finally, a safe path (hypothesis) with high confidence is selected; if all paths
are classified as unsafe, then no path is selected. The algorithm terminates once a path is confirmed safe or all paths are confirmed unsafe.



\section{Background}

\subsection{RRT-Based Non-Holonomic Planning over a Point Cloud}
\label{subsec:background_rrt}

We build upon an existing rapidly-exploring random tree (RRT) \cite{RRT} based planner for finding kinematically feasible
trajectories over non-planar point cloud environments~\cite{krusi2017driving}.
6D robot poses are expressed by transformation matrices belonging to the Special
Euclidean Group SE(3). A matrix $\mathbf{T}_{\text{MR}}$
specifies the position and orientation of a robot-fixed coordinate frame R
expressed in a given reference map frame M.
\cite{krusi2017driving} considers the following planning problem: given start
and goal poses $\mathbf{T}_{\text{MS}}$, $\mathbf{T}_{\text{MG}}$, and a point
cloud $\mathcal{M} = \{\mathbf{m}^i\}$ with $\mathbf{m}^i \in \mathbb{R}^3 $,
compute a connecting trajectory $\pi: \mathbb{R}_{>0} \rightarrow \text{SE(3)}$.
The trajectory has to satisfy a number of constraints up to a given degree of
approximation: contact with the terrain surface, static traversability (e.g.,
bounded roll and pitch angles), and kinematic constraints -- including bounded
continuous curvature. Trajectories are represented as piecewise continuous
functions in the 6D space of robot poses, and are specified by a
sequence of nodes $\widehat{\pi}=[\mathcal{N}^k]$, where each $\mathcal{N}^k$ is
a tuple $(\mathbf{T}_{\text{M}\text{R}^{k}}, \tau^{k}, \mathbf{w}^k,
\kappa^{k})$. Here, $\mathbf{T}_{\text{M}\text{R}^{k}}$ is a 6D pose attached to
the terrain surface, $\tau^k \in [0,1]$ is the associated static traversability
value, $\mathbf{w}^k$ is a parameter vector specifying a short {\em planar}
trajectory segment connecting $\mathbf{T}_{\text{M}\text{R}^{k}}$ to the next
pose in the sequence, and $\kappa^{k}$ is the curvature at the beginning of the
trajectory segment. $\mathbf{w}^k$ specifies a trajectory segment as a 
cubic curvature polynomial~\cite{nagy} evolving along the planar patch defined by the 
$xy$ plane of the coordinate frame $\text{R}$ attached to $\mathbf{T}_{\text{M}\text{R}^{k}}$. 
The end point of such a trajectory segment gives the subsequent pose $\mathbf{T}_{\text{M}\text{R}^{k + 1}}$ through 
a projection on the terrain surface via $f: (\mathcal{M}, \mathbf{T}_{\text{M}\text{R}}) \mapsto \mathbf{T}_{\text{M}\text{R}}$; 
$f$ queries $\mathcal{M}$ for the $K$ nearest-neighbors of the end point, which can be thought of as the points the 
robot will lie on at $\mathbf{T}_{\text{M}\text{R}^{k + 1}}$ ($K$ depends on the size of the robot and point cloud density). 
We use $\phi(\mathcal{N}^{k + 1})$ to denote such points. 


Leveraging the above trajectory representation,~\cite{krusi2017driving} proposes
to define a small set of {\em motion primitives} (short trajectory segments) and
use them to grow two RRTs, 
one from the start pose and one from
  the goal pose, and iteratively try to connect them. Each new pose is associated with a node $\mathcal{N}^k$, which
  is accepted in the tree only if $\tau^k > 0$. 
\cite{krusi2017driving} also proposes a
  technique to derive a better trajectory (in terms of smoothness and distance)
  starting from an initial one. This second optimization stage is not explicitly
  considered in this work, because it is easily generalized and applied to all
  ``safe'' trajectories according to our method.

\subsection{Segmentation with Bayesian Neural Networks }
\label{subsec:bayesian_neural_nets}


Although segmentation networks for 3D point clouds exist \cite{pointnet,
pointnet++, zhou2018voxelnet, qi2018frustum, wang2018dynamic}, 3D data
repositories are focused on object recognition and part segmentation (e.g.,
\cite{shapenet}) or only contain a small number of scenes \cite{semantic3d2}. In
contrast, existing large-scale image segmentation datasets \cite{cocopanoptic,
cocostuff, ade20k, minc} contain varied surfaces and obstacles in diverse real-world
outdoor scenes.  Therefore, we leverage a state-of-the-art image segmentation
network \cite{deeplabv3plus2018} (Section \ref{subsec:pred}), and update the
point cloud environment from per-pixel image segmentations (Section
\ref{subsec:labels}). Furthermore, we augment the network to estimate output
uncertainty, allowing uncertainty in surface predictions to be embedded in the
point cloud. Uncertainty in surface type is used to guide path-safety
evaluation.


At inference time, forward passes with active dropout layers can be interpreted
as an approximation of the posterior distribution of model weights of a neural network
~\cite{Gal2016Dropout, KendallUncert}. The uncertainty
of predictions are computed by taking the sample standard deviation across
multiple forward passes. For each pixel $(i,j)_X$ of image $X$, the mean
softmax vector over $T$ forward passes is:

\vspace{-0.15in}

\begin{small}
\begin{equation} \label{eq:pred}
  \mathbf{p}^{(i,j)_X} = \frac{1}{T} \sum\limits_{t=1}^T
  \mathbf{s}_t^{(i,j)_X}(y|X)
\end{equation}
\end{small}

\vspace{-0.1in}
\noindent where $\mathbf{s}^{(i,j)_X}(y|X) \in \mathbb{R}^{C}$ is the softmax output of the
network. The corresponding uncertainty vector $\boldsymbol{\sigma}^{(i,j)_X}$ on $\mathbf{p}^{(i,j)_X}$ is:

\begin{small}
\begin{equation} \label{eq:uncert}
  \boldsymbol{\sigma}^{(i,j)_X} = \sqrt{\frac{\sum\limits_{t=1}^T
  \big(\mathbf{s}_t^{(i,j)_X}(y|X) -
  \mathbf{p}^{(i,j)_X}\big)^2}{T - 1}}\\
\end{equation}
\end{small}

\vspace{-0.05in}

In our framework, image $X$ corresponds to a view with known camera parameters.
$\mathbf{p}^{(i,j)_X}$ and $\boldsymbol{\sigma}^{(i,j)_X}$ are combined with
existing measurements for each point $\mathbf{m} \in \mathcal{M}$ that
maps to pixel $(i,j)_X$ (Section \ref{subsec:labels}).

\section{Approach Overview}
\label{sec:problem_setting}

\begin{figure*}[tp]
  \centering

\smallskip
  \stackunder[7pt]{\includegraphics[width=2.3in]{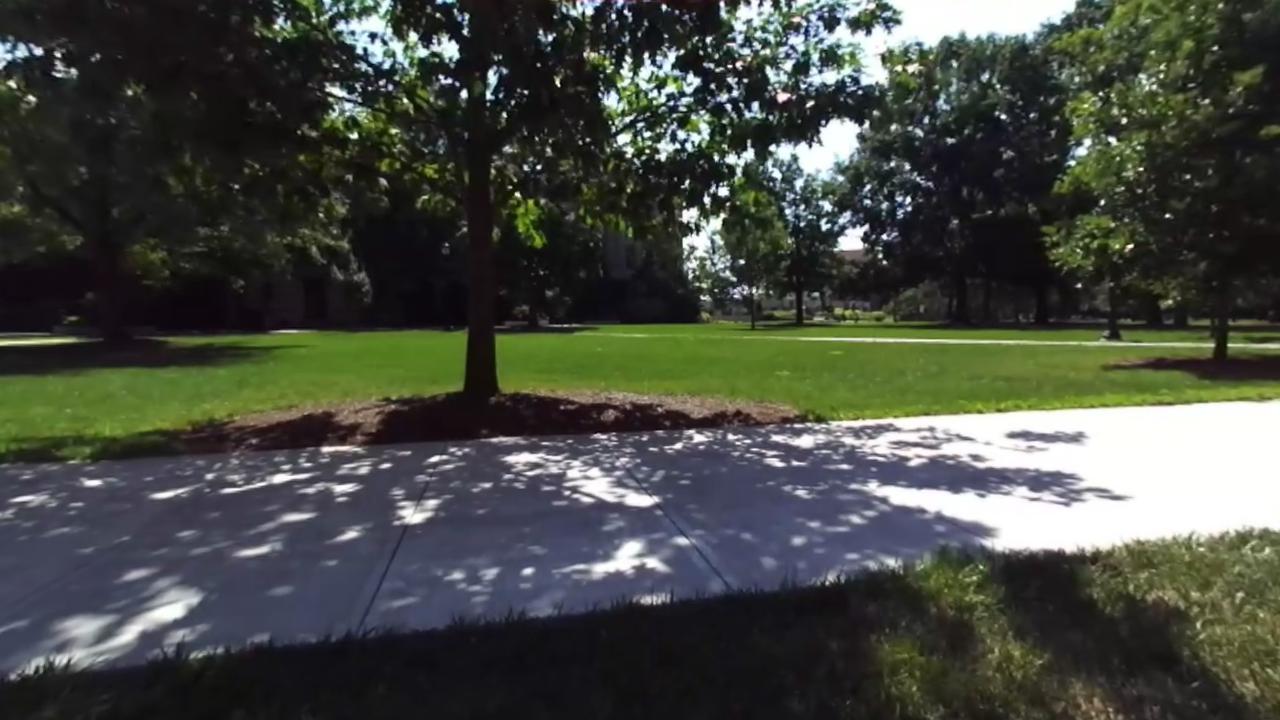}}{(a)}
\stackunder[7pt]{\includegraphics[width=2.3in]{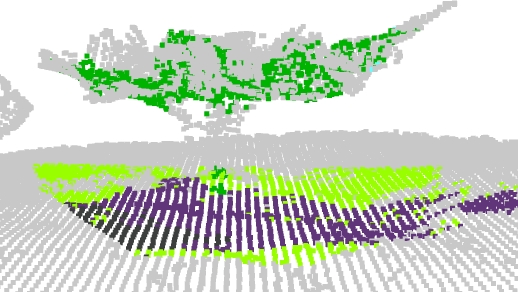}}{(b)}
\stackunder[7pt]{\includegraphics[width=2.3in]{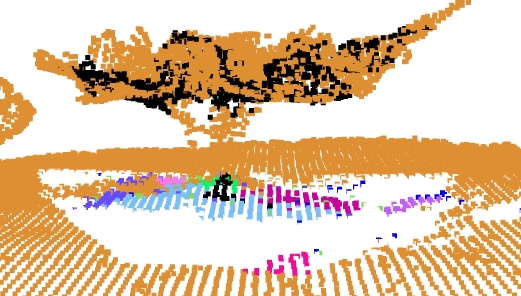}}{(c)}
\caption{An example point cloud. (a) Image view of a portion of the environment. (b) Point
  cloud colored with the most likely class predicted from image (a) (bright green: ``grass''; dark green: ``tree''; purple: ``sidewalk''; dark grey: ``road''; light grey: no information available). All the classes except ``tree'' belong to the set $S$. The region around the tree is actually mulch/woodchips, which should be classified as ``dirt'' (belonging to $U$). (c) Point cloud colored to show safe (white), unsafe (black), unclear regions $\mathcal{R}$ partitioned according to most likely class (random colors), and points not associated with any class (orange). 
  \label{fig:setup}}

\vspace{-0.2in}

\end{figure*}

The planner presented in Section~\ref{subsec:background_rrt} does not leverage critical semantic
information about terrain types. In this work, we assume an initial
point cloud is given in the form $\mathcal{M} = \{\mathbf{e}^i\}$, where each
element $\mathbf{e}_i$ is a tuple $(\mathbf{m}^i, \mathbf{p}^i, \bm{\sigma}^i)$.
Here, $\mathbf{m}^i \in \mathbb{R}^3$ as before; $\mathbf{p}^i=(p_1^i, p_2^i,
\ldots, p_C^i)$ is a vector specifying the probabilities that the point belongs
to each one of the $C$ possible semantic classes (gravel, water, etc.);
$\bm{\sigma}^i=(\sigma_1^i, \sigma_2^i, \ldots, \sigma_C^i)$ is a vector
specifiying the uncertainties of $\mathbf{p}^i$, as discussed in
Section~\ref{subsec:bayesian_neural_nets}. Points are initialized with uniform
semantic probabilities and maximum uncertainties.  Updating the semantic point
cloud is discussed in Section \ref{subsec:labels}.  Fig.~\ref{fig:setup}(b)
shows a pointcloud obtained in a real (ambiguous) environment, where each point $\mathbf{m}^i$ is associated
with the color corresponding to the class label $j$ whose $p_j^i$ is maximum. 

We assume the semantic classes have been partitioned into two sets: the safe
set $S$ (e.g., gravel, grass) and the unsafe set $U$ (e.g., water, snow). For each
point $\mathbf{m}^i$, the points are defined as $p_S^i=\sum_{j \in S}p^i_j$, $p_U^i=1 - p_S^i = \sum_{j \in
U}p^i_j$, and $\sigma^i = \min(\sqrt{\sum_{j \in S}\sigma_j^{i2}},\sqrt{\sum_{j
\in U}\sigma_j^{i2}})$. Each point is then classified as:

\begin{itemize}
\item {\bf safe} if $p_S^i - w_{\sigma}\sigma^i \geq \theta_s$;

\item {\bf unsafe} if $p_U^i - w_{\sigma}\sigma^i \geq \theta_u$;

\item {\bf unclear} otherwise.
\end{itemize}


\noindent Intuitively, this implies that points are safe/unsafe given high
probability ($p_S^i$, $p_U^i$) and low uncertainty ($\sigma^i$), and unclear otherwise. 
$w_{\sigma}$, $\theta_s$, $\theta_u$ are defined by the mission planner (with $1 - \theta_s < \theta_u$). We use
$\mathcal{M}_{\textrm{safe}}$, $\mathcal{M}_{\textrm{unsafe}}$, $\mathcal{M}_{\textrm{unclear}}$ to denote the
partition of $\mathcal{M}$ obtained from the above classification. Note that a point is labeled safe (unsafe) even with large uncertainty on $p_S^i$ ($p_U^i$), provided there is
small uncertainty on $p_U^i$ ($p_S^i$). This is captured by the $\min$ in the
definition of $\sigma^i$. For example, the network may be
uncertain between gravel and grass (both safe), but it is sure that the point is
neither water nor snow (both unsafe). 

Consider now a trajectory
$\widehat{\pi}=[\mathcal{N}^k]$, and recall that $\phi(\mathcal{N}^k)$ denotes
the set of points on which the robot lies when at pose
$\mathbf{T}_{\text{M}\text{R}^{k}}$. Depending on the semantic information
initially available, it might be very difficult --if not impossible-- to
immediately find a trajectory whose node points $\phi(\mathcal{N}^k)$ all belong
to $\mathcal{M}_{\textrm{safe}}$. DeepSemanticHPPC works in two stages:
\begin{itemize}
\item[1)] {\bf Compute a set of candidate paths} traversing different {\em unclear regions}, and

\item[2)] {\bf Reduce the uncertainty of such paths} by taking {\em new views}
  in the proximity of the robot's starting position of the {\em most promising path}.
\end{itemize} 

We relax the path planning problem in a natural way -- instead of reaching a
specific goal pose, we require the robot to reach a goal pose region $G$ defined
around $\mathbf{T}_{\text{MG}}$. 
Then, the points in $\mathcal{M}_{\textrm{unclear}}$ are organized into a set $\mathcal{R}$
of {\em unclear regions}. To build the set $\mathcal{R}$, we use the following
two-stage clustering process:
first, DBSCAN~\cite{dbscan} performs a large-scale clustering of the
points in $\mathcal{M}_{\textrm{unclear}}$, obtaining a set of large unclear regions
$\widehat{\mathcal{R}}$. Then, the points of each $\hat{r} \in
\widehat{\mathcal{R}}$ are further partitioned according to their most likely
class (treating the points not associated with any prediction as belonging to a
special class); DBSCAN is called again on each partition.
Fig.~\ref{fig:setup}(c) shows the result of this process on our example with $\theta_s=0.9, \theta_u=0.3, w_{\sigma}=3$. 

The remaining task is to compute a set of candidate paths from $\mathbf{T}_{\text{MS}}$ to $G$.
Section~\ref{subsec:rrt} presents a variant of a standard RRT algorithm which constructs multiple hypothesis for the safest path, traversing different unclear
regions. These paths are stored in the directed graph $G=(V,A)$, where
each $v \in V$ is associated with a potential trajectory node $\mathcal{N}^v$
and each $a \in A$ represents the existence of a short trajectory segment
connecting two poses.
\ The cost for each node is based on how far it is from satisfying our
safety constraint:  $\bar{p}^v= \frac{1}{|\phi(\mathcal{N}^v)|}\sum_{i \in
        \phi(\mathcal{N}^v)}\min(1,\max(0, \theta_s - p_S^i +
        w_{\sigma}\sigma^i))$ for for each $v \in V$.
However, if the node region sufficiently intersects with an unsafe region, the cost is
infinite; and if the node lies entirely in a safe region, the cost is zero.

Summarizing, $c(v)$ is defined as:

\begin{small}
\begin{equation} \label{eq:cost}
c(v) = \begin{cases}
        0 & \text{ if } |\mathcal{M}_{\textrm{safe}} \cap \phi(\mathcal{N}^v)| = |\phi(\mathcal{N}^v)| \\
        \infty & \text{ if } |\mathcal{M}_{\textrm{unsafe}} \cap \phi(\mathcal{N}^v)| \geq \phi_v \\
        \bar{p}^v & \text{ otherwise},
        \end{cases}
\end{equation}
\end{small}

\vspace{-0.12in}

\noindent where $\phi_v$ is a user-defined threshold. The first condition can be relaxed for the nodes in close proximity of the starting pose. Although a vertex with
infinite cost is never obtained when the candidate paths are initially computed,
its cost might tend to infinity when additional views are taken during the NBV stage
(Section~\ref{subsec:nbv}). A predefined number of NBV iterations are run. At
each iteration, the $m$ most promising (shortest) paths according to the above
cost function are computed by a $k$-shortest paths algorithm (we use Yen's~\cite{yen1971finding}). 
The associated vertices $v$ such that $0 < c(v)
< \infty$ are then considered in the function that computes the best additional
view. The value of $m$ is also decided by the mission planner: $m=1$
corresponds to an aggressive setting, while $m > 1$ is preferred given a large
temporal budget for taking additional views.

\section{Technical Details}

\subsection{Predicting Semantic Labels}
\label{subsec:pred}


We curate a segmentation dataset for outdoor navigation in unstructured
environments from the existing large-scale COCO panoptic dataset
\cite{cocopanoptic}.  First, images of unstructured outdoor scenes are selected
using a Places365 \cite{places} classifier. An image is kept if (a) the
classifier's top one (highest) prediction is an unstructured outdoor category
with $>50\%$ probability, or (b) two or more of the top five predictions are
unstructured outdoor categories. Second, the 133 categories in COCO panoptic are
merged: (a) all outdoor terrains (e.g., grass, dirt, snow, pavement) are
retained; (b) obstacles are merged into four categories: fixed obstacles (e.g.,
buildings), moving human-made obstacles (e.g., vehicles), humans, and animals;
(c) all indoor categories are removed. Our final dataset consists of 34K
training images with 22 categories.  Our navigation segmentation categories
(from COCO panoptic), and filtered list of COCO panoptic images are at:
{\color{red}\texttt{https://deepsemantichppc.github.io}}

For our network architecture, we use DeepLabv3+\cite{deeplabv3plus2018} with
Xception65~\cite{chollet2017xception} backbone augmented with dropout in the
middle and exit flow blocks for semantic segmentations. 
%
At inference time, 50 forward passes are used to predict semantics and
uncertainties (Eqs. \eqref{eq:pred}-\eqref{eq:uncert}).

\subsection{Associating Semantic Labels to a Point Cloud}
\label{subsec:labels}
Given a viewpoint with known pose and camera intrinsics, image
segmentation probabilities and uncertainties are mapped to the point cloud. To map
pixel $(i,j)_X$:
\begin{enumerate}
  \item Estimate depth map $\mathbf{D}_X$ for view X.
  \item Each pixel is backprojected to a single point
    corresponding to the center of the projected pixel. This approximation does
    not hold for pixels with very large depths, but typically produces good results. Backprojected pixel point $\mathbf{\tilde{m}}^{(i,j)_X}$ is
    computed as:

    \vspace{-0.15in}
    \begin{small}
      \begin{equation}
      \mathbf{\tilde{m}}^{(i,j)_X} = \mathbf{P}_X [\mathbf{D}_X^{(i,j)}\mathbb{I}_{3\times3} | [0~0~1]^T]^T \mathbf{K}_X^{-1}
    {[i~j~1]^T}
    \end{equation}
    \end{small}
    \vspace{-0.15in}


    where $\mathbf{K}_X$, $\mathbf{P}_X$ are intrinsic and pose matrices.
  \item Each backprojected point is merged with its nearest neighbor
    $\mathbf{m}^\textrm{nn}$ in the
    point cloud $\mathcal{M}$ within threshold distance $R$. If a backprojected point does
    not have a neighbor within the threshold, the point is discarded.


    \item $\mathbf{p}^{(i,j)_X}$ and
  $\boldsymbol{\sigma}^{(i,j)_X}$ are merged with existing measurements for
  point $\mathbf{m}^{\textrm{nn}}$. The combined measurement is the best linear
  unbiased estimator under the simplifying assumption that the per-class
  predictions are independent.
  Given a set of $K$ measurements
  $\{(\mathbf{p}^{(k)}, \boldsymbol{\sigma}^{(k)})\}$, the combined measurement
 $(\mathbf{\tilde{p}}, \tilde{\boldsymbol{\sigma}})$
  for class $c$ is:

  \vspace{-0.1in}

\begin{small}
\begin{align} \label{eq:todo}
  \nonumber &\Bigg(\tilde{p}_c = \dfrac{1}{Z}\sum\limits_{k=1}^K {w}^{(k)}{p}^{(k)}_c
  \text{\quad,\quad} \tilde{\sigma}_c = \sqrt{\sum\limits_{k=1}^K
  ({w}^{(k)})^2 ({\sigma}^{(k)})^{2}} ~\Bigg)\\
  &\text{ where } {w}^{(k)}_c = \dfrac{(\sigma_c^{(k)})^{-2}}{\sum\limits_{c'
  \in C}
  (\sigma_{c'}^{(k)})^{-2}} \quad, \quad Z \text{ s.t. } \sum\limits_{c'
  \in C} \tilde{p}_{c'}=1 &&
\end{align}
\end{small}
\end{enumerate}

%

\subsection{RRT-Based Multi-hypothesis Planner}
\label{subsec:rrt}

\begin{figure}[tp]
\centering

\vspace{2mm}
\smallskip
    {\includegraphics[width=0.6\columnwidth]{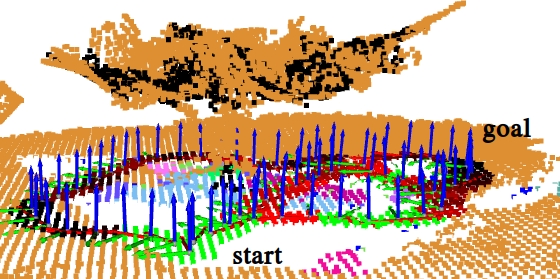}}
	\caption{An example graph $G=(V,A)$, with poses. Vertices with $c(v)=0$ are in green, while vertices with $0 < c(v) \le 1$ are in red (the darker, the closer to 1). The blue, green, and red axes indicate the pose of the robot.
    \label{fig:multipath_example}}
    \vspace{-0.25in}
\end{figure}

The algorithm to compute $G=(V,A)$ 
starts by building an initial RRT from a root vertex $v_s$ associated with the projected start pose
$\mathbf{T}_{\text{M}\text{S}}$ via a predefined set of motion primitives. Instead of building two RRTs as is
done in~\cite{krusi2017driving}, we build a single RRT from the start pose and bias the sampling to the point 
that is closest to the projected ideal goal pose. Sampling is performed on the points in $\mathcal{M}$ not lying
in the {\em forbidden points set} $\mathcal{F}$, which is initialized as $\mathcal{F} \leftarrow \mathcal{M}_{\textrm{unsafe}}$. Once the first path $\widehat{\pi}=[\mathcal{N}^v]$ is found, the algorithm examines which regions
in $\mathcal{R}$ are traversed by the vertices of $\widehat{\pi}$ by checking their intersections with each set of points $\phi(\mathcal{N}^v)$. Except for regions containing points belonging to the $K$-nearest neighbors of $\mathbf{T}_{\text{M}\text{S}}$ and $\mathbf{T}_{\text{M}\text{G}}$, all regions traversed by
 $\widehat{\pi}$ are placed into the {\em removal candidates set} $\mathcal{C}$. A heuristic is then used to decide which region(s) should be removed from subsequent planning stages. In this work, we simply remove the largest region $\hat{c}$, and $\mathcal{F}$ is updated as $\mathcal{F} \leftarrow \mathcal{F} \cup \hat{c}$. When $\hat{c}$ is removed, the RRT vertices lying on it, including those of $\hat{\pi}$, are removed from the RRT. The algorithm then proceeds to optimize and find a new path to the goal, by continuously expanding the RRT component containing $v_s$. This time, however, the algorithm also tries to connect (by computing {\em ad hoc} trajectory segments) new vertices to those that were disconnected from the RRT due to the removal of $\hat{c}$. When a new path is found, the process repeats. If at any iteration, the algorithm finds a path only contained in the regions of $\mathbf{T}_{\text{M}\text{S}}$ and $\mathbf{T}_{\text{M}\text{G}}$, it can be restarted with a different random seed. The two graphs are merged at a later stage. Fig.~\ref{fig:multipath_example} shows an example multi-hypothesis graph computed on the example of Section~\ref{sec:problem_setting} (Fig. \ref{fig:setup}).

If any path computed on the multi-hypothesis graph $G=(V,A)$ has zero cost, the robot starts to follow that path since all the underlying points lie in $\mathcal{M}_{\textrm{safe}}$. Otherwise, the robot enters the NBV stage described below.

\subsection{Next-Best-View (NBV) Planning}
\label{subsec:nbv}

A set of $n$ viable NBV poses $\mathbf{T}_{\text{viable}} =
\{\mathbf{T}_{\text{MV}_1}, \ldots, \mathbf{T}_{\text{MV}_n}\}$ is computed by
growing a RRT starting from $\mathbf{T}_{\text{MS}}$.  The candidate poses
$\mathbf{T}_{\text{viable}}$ are a subsample of the RRT vertices. All points
lying within $\mathcal{M}_{\textrm{unclear}}$ are treated as {\em unsafe}, and
sampling is performed on the safe points lying within a given radius $r$ from
$\mathbf{T}_{\text{MS}}$. The robot should not travel too far to take a new
view; otherwise it is more appropriate to follow the most promising path of
$G=(V,A)$. 

A reward $J(\mathbf{T}_{\text{MV}_j})$ is calculated for each candidate pose
$\mathbf{T}_{\text{MV}_j}$. The NBV pose $\mathbf{T}_{\text{NBV}}$ is
selected by picking the pose with the highest value of $J$ that also allows a
safe path back to a safe pose from which all the paths can be followed.  

\begin{small}
\begin{equation}
J(\mathbf{T}_{\text{MV}_j}) = \beta_{d}  D + \beta_{\gamma}  \gamma + \beta_{\text{vis}}  N_{\text{vis}} + \beta_{Q}  \overline{Q},
\label{eq:nbv}
\end{equation}
\end{small}

\vspace{-0.2in}

\noindent where $D$ is a distance metric, $\gamma$ is a viewing angle metric, $N_{\text{vis}}$ is the
number of visible vertices from $\mathbf{T}_{\text{MV}_j}$, and $\overline{Q}$
is the average information gain over visible vertices from
$\mathbf{T}_{\text{MV}_j}$. The weights $\beta_{d}, \beta_{\gamma},
\beta_{\text{vis}}, \beta_{Q}$ sum to 1 and $D$, $\gamma$, $N_{\text{vis}}$, and
$\overline{Q}$ are normalized. Distance and change in viewing angle are
used based on the assumption that closeness and view diversity will reduce
vertex uncertainty. $N_{\text{vis}}$ puts more weight on candidate poses which
have higher chances of reducing the uncertainty of multiple segments of the
multipath graph $G=(V,A)$. $\overline{Q}$ represents the expected reduction in
uncertainty in the graph vertices given $\mathbf{T}_{\text{MV}_j}$. Each of
these components are defined as follows.

Begin by defining the set of vertices $v \in V_{\text{NBV}}$, where
$V_{\text{NBV}}$ is the set of unclear vertices belonging to the $m$ most
promising paths (see the end of Section~\ref{sec:problem_setting}). For each
candidate pose $\mathbf{T}_{\text{MV}_j} \in \mathbf{T}_{\text{viable}}$, only
vertices visible from $\mathbf{T}_{\text{MV}_j}$ are considered in calculating
the reward. Visible vertices are defined as vertices which occupy greater than a
predefined number of pixels in the image plane rendering of the point cloud from
$\mathbf{T}_{\text{MV}_j}$. Visible vertices are added to the set $v \in
V_{\text{vis},j}$, where $V_{\text{vis},j} \in V_{\text{NBV}}$. 

To calculate $D$, the distances from $\mathbf{T}_{\text{MV}_j}$ to $v \in
V_{\text{vis},j}$ are normalized. Since a lower distance should correspond to a
higher reward, we subtract the normalized distances from 1. To calculate
$\gamma$: the set of negative cosine distances of the angle between
$\mathbf{T}_{\text{MS}}$ and $\mathbf{T}_{\text{MV}_j}$ to $v \in
V_{\text{vis},j}$ are used. $N_{\text{vis}}$ is the size of
$V_{\text{vis},j}$.

The information gain metric $Q(v)$ is calculated for each $v \in
V_{\text{NBV}}$. $Q(v)$ represents the expected reduction in uncertainty for
each $v \in V_{\text{NBV}}$, and is a function of the visibility and uncertainty
of the points lying in $\phi(\mathcal{N}^v)$. 

The visibility $I(v, \mathbf{T}_{\text{MV}_{j}})$ of a vertex $v \in V_{\text{NBV}}$, given a
candidate pose $\mathbf{T}_{\text{MV}_{j}}$, is the pixel coverage of $\phi(\mathcal{N}^v)$ in
the rendered image plane of $\mathbf{T}_{\text{MV}_{j}}$. Per-point bounding squares of size equal to
half the point cloud resolution are used
to compute occlusions and pixel coverage for surface points. The number of
pixels that $\phi(\mathcal{N}^v)$ occupies in the rendering is the predicted
visibility $I(v, \mathbf{T}_{\text{MV}_{j}})$ of $v$ at $\mathbf{T}_{\text{MV}_{j}}$.

The uncertainty $\sigma(v)$ of a vertex $v \in V_{\text{NBV}}$ is the average
sum of the uncertainties of the points in $\phi(\mathcal{N}^v)$:

\vspace{-0.1in}

\begin{small}
\begin{equation}
\sigma(v) = \frac{1}{|\phi(\mathcal{N}^v)|} \sum_{i \in \phi(\mathcal{N}^v)} \sum_{j = 1}^{C}\sigma_i^j
\end{equation}
\end{small}

\vspace{-0.1in}

\noindent The information gain metric $Q(v)$ of vertex $v \in V_{\text{NBV}}$ can be formally written as

\begin{small}
\begin{equation}
Q(v) =  \alpha_{i}  I(v, T_{MV_{j}}) + \alpha_{\sigma}  \sigma(v),
\end{equation}
\end{small}

\vspace{-0.2in}

\noindent where $\alpha_{I}, \alpha_{\sigma}$ are weights that sum to 1 and
$I(v, \mathbf{T}_{\text{MV}_{j}}) \textrm{ and } \sigma(v)$ are normalized.

\section{Validation}
\subsection{Validation Scenes and Overview}




We implement our full pipeline in the AirSim simulator \cite{airsim}.
However, preliminary experiments show that the BNN trained on the real-world
dataset (Section~\ref{subsec:pred}) performs poorly on the synthetic
AirSim environment. The full pipeline with a simpler BNN trained on the AirSim
environment and corresponding experimental results are shown in the accompanying
video.

\begin{figure}[tp]
\centering 

\smallskip
{\includegraphics[width=0.9\columnwidth]{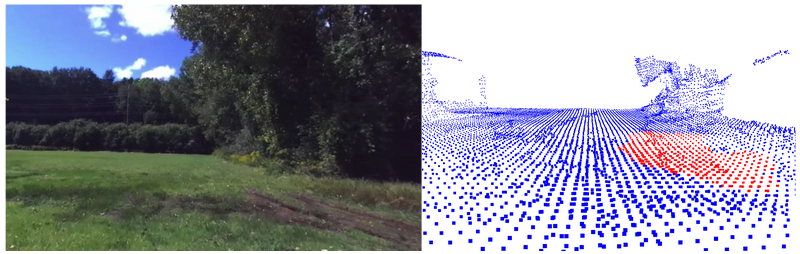}}
\caption{\textbf{Left}: Image from Cass Park in Ithaca. There are multiple
different terrains in the scene including grass, mud, and water.
\textbf{Right}: Annotated safe (blue) and unsafe (red) regions for the Cass Park point cloud.
  \label{fig:Cass}}
  \vspace{-0.25in}
\end{figure}


For real world validation, we collect data of two different scenes using the
ZED stereo camera from Stereolabs. The first scene is next to the Mann Library
in Cornell University (Fig. \ref{fig:setup}(a)) and the second scene is at Cass
Park in Ithaca (Fig. \ref{fig:Cass}). These scenes are selected due to their
varying (but common) terrain types. The scenes are representative of common
unstructured outdoor environments. The
ZED camera API is used to extract depth maps and generate point cloud
reconstructions of the scenes.
For these scenes, we heuristically select a set of candidate NBV poses instead
of growing a RRT from $\mathbf{T}_{\text{MS}}$, and assume a path between these
poses and the start pose exists. Candidate NBV poses are
selected to be oriented in the general direction of the goal while covering a wide
range of the scene. Our method (and baseline methods) choose NBVs from the
set of candidate NBV poses.

\subsection{NBV evaluation}

\begin{figure}[tp]
\centering

\smallskip
    {\includegraphics[width=1.0\columnwidth]{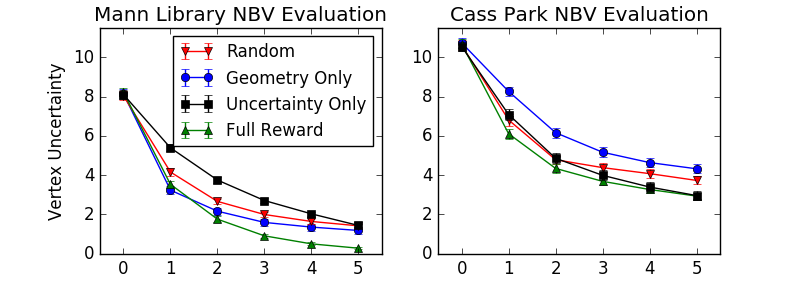}}
    \vspace{-3mm}
	\caption{Change in uncertainty of path vertices (y-axis) as the number of NBV
	measurements increase (x-axis). 
    \label{fig:nbv_eval}}
    \vspace{-0.25in}

\end{figure}

To evaluate the performance of the NBV function, we examine the change in
uncertainty of the path vertices as the number of NBVs increases.  The complete
NBV reward function (Eq. \ref{eq:nbv}) is compared
against: (a) random selection, (b) geometry-only reward, and (c) uncertainty-only
reward. For the geometry-only NBV reward, we set $\{\beta_{Q}\}$ to zero, and for the
uncertainty-only NBV reward, we set $\{\beta_{d}, \beta_{\gamma},
\beta_{\textrm{vis}}\}$ to zero. The change in uncertainties summed across all
the classes and points for each path vertex averaged over 500 trials is shown in
Fig.  \ref{fig:nbv_eval}. In our experiments, the full NBV reward weights are set
as follows: $\{\beta_{d}=0.4, \beta_{\gamma}=0.05, \beta_{\text{vis}}=0.25,
\beta_{Q}=0.3, \alpha_{I}=0.5, \alpha_{\sigma}=0.5\}$ for the Mann Library
scene, and $\{\beta_{d}=0.15, \beta_{\gamma}=0.05, \beta_{\textrm{vis}}=0.2,
\beta_{Q}=0.6, \alpha_{I}=0.3, \alpha_{\sigma}=0.7\}$ for the Cass Park scene. A
higher weight is assigned to uncertainty terms for the Cass scene because the boundaries
between surface types (e.g., water, mud, grass) are more ambiguous than in the
Mann scene.

For both scenes, the full NBV reward function consistently achieves the lowest
uncertainty with 2 or more NBVs (Fig. \ref{fig:nbv_eval}). This illustrates the
importance of both geometry and uncertainty terms in the reward function. In the
Mann scene, the baseline reward functions converge to a higher uncertainty than
the full reward function. Due to the small size of the scene and the nature of all
candidate NBVs being oriented towards the goal pose, random selection performs
quite well.  It initially outperforms uncertainty-only which does not take into
account point visibility or viewpoint diversity, while random sampling
implicitly selects diverse views. In the Cass scene, the baseline reward functions
converge to a higher uncertainty than the full reward function, except for
uncertainty-only which converges to the same point as the full reward function. The overall
uncertainty in the BNN predictions are much higher at Cass park, so heavily
weighting the uncertainty in the reward function performs well. In the Mann scene,
geometry-only outperforms uncertainty-only, whereas the opposite result holds
for the Cass scene. Mann has less inherent ambiguity so geometry terms are more
important, while Cass has more ambiguous regions so uncertainty terms
are more important.

\subsection{Path Safety Evaluation}

To evaluate the real world application of DeepSemanticHPPC, we study the safety
of selected paths. We annotate a point cloud (Fig. \ref{fig:Cass}
\textbf{right}) with MeshLab \cite{meshlab} into ground truth safe and unsafe
regions. For Mann library mulch (dirt) is labeled as unsafe, and for Cass Park
mud (dirt) and water are labeled as unsafe. Any path which contains a vertex
that overlaps with the unsafe region with over $N_{\text{unsafe}}$ points is
unsafe. We set $N_{\text{unsafe}} = 4$. Two baselines are considered: (a) B1:
planning without semantic information (based on \cite{krusi2017driving}) and (b)
B2: planning with semantic information from a single initial view without taking any
NBV measurements to reduce path uncertainty. We also study the performance of
DeepSemanticHPPC as the number of NBVs increase. 


\begin{table}[tp]
\centering

\vspace{2mm}
\smallskip
\begin{tabular}{|c|c|c|c|c|c|c|c|}
\hline
\textbf{\!\!\textit{Mann}\!\!}   & \textbf{B1} & \textbf{B2} & \textbf{1N} & \textbf{2N} & \textbf{3N} & \textbf{4N} & \textbf{5N} \\ \hline

\textbf{\!\!Safe \%\!\!} & \!\!0\!\!         & \!\!23.4\!\!         &
\!\!81.6\!\!            & \!\!79.8\!\!
&\!\!81.4\!\!
& \!\!83.6\!\!           & \!\!\underline{\textbf{86.0}}\!\!            \\ \hline
\textbf{\!\!Unsafe \%\!\!} & \!\!100\!\!         & \!\!76.6\!\!         &
\!\!18.4\!\!            & \!\!15.4\!\! 
&\!\!11.2\!\!
& \!\!6.6 \!\!           & \!\!\underline{\textbf{3.8}}\!\!            \\ \hline

\textbf{\!\!CS \%\!\!}    & N/A         & N/A         & 0             & 0
& \!\!4.0\!\!
& \!\!18.0\!\!          & \!\!28.0\!\!        \\ \hline
\textbf{\!\!CN \%\!\!}    & N/A         & N/A         & 0             &
\!\!4.8\!\!
& \!\!7.4\!\!
& \!\!9.8\!\!            & \!\!10.2\!\!              \\ \hline
\textbf{\!\!\textit{Cass}\!\!}   & \textbf{B1} & \textbf{B2} & \textbf{1N} & \textbf{2N} & \textbf{3N} & \textbf{4N} & \textbf{5N} \\ \hline

\textbf{\!\!Safe \%\!\!} & \!\!13.2\!\!         & \!\!33.4\!\!       &
\!\!54.0\!\!           & \!\!57.4\!\!            &
\!\!57.4\!\!            & \!\!59.0\!\!            & \!\!\underline{\textbf{59.2}}\!\!            \\ \hline
\textbf{\!\!Unsafe \%\!\!} & \!\!86.8\!\!         & \!\!66.6\!\!       &
\!\!46.0\!\!           & \!\!41.6\!\!            &
\!\!39.8\!\!            & \!\!37.4\!\!            & \!\!\underline{\textbf{36.6}}\!\!            \\ \hline

\textbf{\!\!CS \%\!\!}    & N/A         & N/A         & 0             & 0             & 0            & 0           & 0           \\ \hline
\textbf{\!\!CN \%\!\!}    & N/A         & N/A         & 0             &
\!\!1.0\!\!
& \!\!2.8\!\!
& \!\!3.6\!\!            & \!\!4.2\!\!            \\ \hline
\end{tabular}
\caption{500 trials of path safety evaluation. The columns are the path planning
  methods used: \textbf{B1} is the planner based on \cite{krusi2017driving},
  \textbf{B2} is the variant of our framework which does not utilize any NBVs,
  \textbf{$X$N} is DeepSemanticHPPC (ours) with $X$ NBVs. The rows are the metrics:
  \textbf{Safe} is the number of trials where a safe path is selected,
  \textbf{Unsafe} is the number of trials where an unsafe path is selected
  (lower is better), \textbf{CS} is the number of trials  where a safe path is
  confirmed, \textbf{CN} is the number of trials where all multipaths  are
  confirmed as unsafe (and no path is selected). 
} \label{table:2}
\vspace{-0.25in}
\end{table}

Table \ref{table:2} shows the path safety results for the Mann Library and Cass
Park scenes over 500 trials. Start and goal poses are varied for each trial within a one meter radius of reference poses. A path is confirmed safe (\textbf{CS}) if it has cost zero. A graph $G = (V,A)$ is confirmed unsafe (\textbf{CN}) if all paths have infinite cost. 
Since both safe and unsafe terrain surfaces can be
geometrically similar, baseline B1 cannot reliably avoid unsafe semantic
regions. 
Baseline B2 performs significantly better than B1 because of the inclusion of
semantic surface types in the planner. However, because the semantic
segmentations can be incorrect, especially in regions with high uncertainty,
this planner still plans over unsafe terrain.

Our DeepSemanticHPPC framework is significantly better than the two baselines.
NBVs allow the planner to discard unsafe paths as semantic uncertainties
decrease.  With just one NBV, the percentage of safe paths taken increases
drastically from 23.4\% to 81.6\% (Mann) and increases from 33.4\% to 54.0\%
(Cass). As NBVs increase, the percentage of safe paths selected
generally increases while the percentage of unsafe paths selected decreases.
With 5 NBVs, 86\% (59.2\%) of paths selected for Mann (Cass) are safe, and only
3.8\% (36.6\%) of paths selected for Mann (Cass) are unsafe. With 5 NBVs,
uncertainty is sufficiently reduced so that in 10.2\% (4.2\%) of trials, all
multipaths are confirmed to be unsafe for Mann (Cass) and no path is selected.
The complexity of the Cass scene is reflected in these results.




\section{Conclusion}
In this paper we present DeepSemanticHPPC, a novel framework for planning in
unstructured outdoor environments while accounting for uncertain terrain types.
Our experiments show DeepSemanticHPPC reduces semantic
uncertainty in planned paths and increases the safety of paths planned in
environments with unsafe terrains. We plan to implement DeepSemanticHPPC on a
robot for physical experiments. Other interesting directions include exploring
the ability to build the point cloud online and incorporating geometric
uncertainties.

\section*{Acknowledgments}
We thank Eric Wu (Cornell) and Hadi AlZayer (Cornell) for insightful discussions and assistance
with preliminary software implementation.

\bibliographystyle{IEEEtran}
\bibliography{paper}

\end{document}